%% file: main.tex
\documentclass[10pt,twocolumn,letterpaper]{article}

\usepackage{cvpr}              

\input{preamble}

\definecolor{cvprblue}{rgb}{0.21,0.49,0.74}
\usepackage[pagebackref,breaklinks,colorlinks,citecolor=cvprblue]{hyperref}

\usepackage[accsupp]{axessibility}  

\usepackage{xcolor, soul}
\sethlcolor{pink}

\def\paperID{105} 
\def\confName{CVPR}
\def\confYear{2024}

\title{Photo-Realistic Image Restoration in the Wild\\ with Controlled Vision-Language Models}

\author{Ziwei Luo$^1$ \, Fredrik K. Gustafsson$^2$ \, Zheng Zhao$^1$ \, Jens Sjölund$^1$ \, Thomas B. Schön$^1$\\
$^1$Uppsala University \,
$^2$Karolinska Institutet\\
{\tt\small \{ziwei.luo,zheng.zhao,jens.sjolund,thomas.schon\}@it.uu.se, fredrik.gustafsson@ki.se}  \\
{\tt\small\url{https://github.com/Algolzw/daclip-uir}}
}

\begin{document}
\maketitle
\input{sec/0_abstract}    
\input{sec/1_intro}
\input{sec/2_related_work}
\input{sec/3_method}

\input{sec/4_experiment}

\input{sec/5_conclusion}

\vspace{5pt}
\noindent \textbf{Acknowledgements}
This research was partially supported by the \emph{Wallenberg AI, Autonomous Systems and Software Program (WASP)} funded by the Knut and Alice Wallenberg Foundation, by the project \emph{Deep Probabilistic Regression -- New Models and Learning Algorithms} (contract number: 2021-04301) funded by the Swedish Research Council, and by the
\emph{Kjell \& M{\"a}rta Beijer Foundation}.
The computations were enabled by the \textit{Berzelius} resource provided by the Knut and Alice Wallenberg Foundation at the National Supercomputer Centre.

{
    \small
    \bibliographystyle{ieeenat_fullname}
    \bibliography{main}
}


\end{document}

%% file: preamble.tex
\usepackage{graphicx}
\usepackage{amsmath}
\usepackage{amssymb}
\usepackage{booktabs}

\usepackage{mathtools}
\usepackage{amsthm}

\usepackage{multirow}

\usepackage[dvipsnames]{xcolor}

\newcommand{\diff}{\mathop{}\!\mathrm{d}}
\newcommand{\expp}{\mathrm{e}}
\newcommand{\cond}{{\;|\;}}

\theoremstyle{plain}

\theoremstyle{definition}

\theoremstyle{remark}

%% file: sec/0_abstract.tex
\begin{abstract}
Though diffusion models have been successfully applied to various image restoration (IR) tasks, their performance is sensitive to the choice of training datasets. Typically, diffusion models trained in specific datasets fail to recover images that have out-of-distribution degradations. To address this problem, this work leverages a capable vision-language model and a synthetic degradation pipeline to learn image restoration in the wild (wild IR). More specifically, all low-quality images are simulated with a synthetic degradation pipeline that contains multiple common degradations such as blur, resize, noise, and JPEG compression. Then we introduce robust training for a degradation-aware CLIP model to extract enriched image content features to assist high-quality image restoration. Our base diffusion model is the image restoration SDE (IR-SDE). Built upon it, we further present a posterior sampling strategy for fast noise-free image generation. We evaluate our model on both synthetic and real-world degradation datasets. Moreover, experiments on the unified image restoration task illustrate that the proposed posterior sampling improves image generation quality for various degradations.
\end{abstract}

%% file: sec/1_intro.tex
\section{Introduction}
\label{sec:intro}


Diffusion models have proven effective for high-quality (HQ) image generation in various image restoration (IR) tasks such as image denoising~\cite{kawar2022denoising,luo2023image,chung2022diffusion}, deblurring~\cite{whang2022deblurring,chen2023hierarchical,ren2023multiscale}, deraining~\cite{ozdenizci2023restoring,yue2023image}, dehazing~\cite{luo2023image,luo2023refusion}, inpainting~\cite{lugmayr2022repaint,saharia2022palette,rombach2022high}, super-resolution~\cite{saharia2022image,li2022srdiff,yue2023resshift}, shadow removal~\cite{luo2023refusion,guo2023shadowdiffusion}, etc. Compared to traditional deep learning-based approaches that directly learn IR models using an $\ell_1$ or $\ell_2$ loss~\cite{chen2022simple,liang2021swinir,zamir2021multi,zamir2020learning} or an adversarial loss~\cite{ledig2017photo,wang2018esrgan,wang2021real}, diffusion models are known for their ability to generate photo-realistic images with a stable training process. However, they are mostly trained on fixed datasets and therefore typically fail to recover high-quality outputs when applied to real-world scenarios with unknown, complex, out-of-distribution degradations~\cite{wang2021real}. 

Although this problem can be alleviated by leveraging large-scale pretrained \textit{Stable Diffusion}~\cite{rombach2022high,podell2023sdxl} weights~\cite{wang2023exploiting,lin2023diffbir,yu2024scaling} and synthetic low-quality (LQ) image generation pipelines~\cite{wang2021real,sahak2023denoising}, it is still challenging to accurately restore real-world images in the wild (i.e., \textit{wild IR}). On the one hand, Stable Diffusion uses an adversarially trained variational autoencoder (VAE) to compress the diffusion to latent space, which is efficient but loses image details in the reconstruction process. Moreover, in practice, the restoration in latent space is unstable and tends to generate color-shifted images~\cite{lin2023diffbir}. On the other hand, most existing works use a fixed degradation pipeline (with different probabilities for each degradation) to generate low-quality images~\cite{wang2021real}, which might be insufficient to represent the complex real-world degradations.

In this work, we aim to perform photo-realistic image restoration with enriched vision-language features that are extracted from a degradation-aware CLIP model (DACLIP~\cite{luo2023controlling}). For scenes encountered in the wild, we assume the image only contains mild, common degradations such as light noise and blur, which can be difficult to represent by text descriptions. We thus add a fidelity loss to reduce the distance between the LQ and HQ image embeddings. Then the enhanced LQ embedding is incorporated into the image restoration networks (such as the U-Net~\cite{ronneberger2015u} in IR-SDE~\cite{luo2023image}) via cross-attention. Inspired by Real-ESRGAN~\cite{wang2021real}, we also propose a new degradation pipeline with a random shuffle strategy to improve the generalization. An optimal posterior sampling strategy is further proposed for IR-SDE to improve its performance. \cref{fig:teaser} shows the comparison of our method with other state-of-the-art wild IR approaches.
In summary, our main contributions are as follows:
\begin{itemize}[leftmargin=.3in]
    \item We present a new synthetic image generation pipeline that employs a random shuffle strategy to simulate complex real-world LQ images.
    \item For degradations in the wild, we modify DACLIP to reduce the embedding distance of LQ-HQ pairs, which enhances LQ features with high-quality information.
    \item We propose a posterior sampling strategy for IR-SDE~\cite{luo2023image} and show that it is the optimal reverse-time path, yielding a better image restoration performance.
    \item Extensive experiments on wild IR and other specific IR tasks demonstrate the effectiveness of each component of our method.
\end{itemize}


%% file: sec/2_related_work.tex
\section{Related Work}
\label{sec:related_work}

\paragraph{Blind Image Restoration}
Image restoration (IR) aims to reconstruct a high-quality (HQ) image from its corrupted counterpart, i.e. from a low-quality (LQ) image with task-specific degradations~\cite{dong2016accelerating,zhang2018residual,li2022d2c,zamir2021multi,zhang2017learning,zamir2020learning,lian2023kernel,zhang2021plug}. Most learning-based approaches directly train neural networks with an $\ell_1$/$\ell_2$ loss on HQ-LQ image pairs, which is effective but often overfit on specific degradations~\cite{zhang2021designing,wang2021real,yu2024scaling}. Thus the blind IR approach is proposed and has gained growing attention in addressing complex real-world degradations. BSRGAN~\cite{zhang2021designing} is the pioneering work that designs a practical degradation model for blind super-resolution, and Real-ESRGAN~\cite{wang2021real} improves it by exploiting a `high-order' degradation pipeline. Most subsequent blind IR methods~\cite{chen2022real,lin2023diffbir} follow their degradation settings but with some improvements in architectures and loss functions. Recently, some works~\cite{li2022all,potlapalli2023promptir,luo2023controlling} further propose to jointly learn different IR tasks using a single model to improve the task generalization, so-called unified image restoration. 

\paragraph{Photo-Realistic Image Restoration}
Starting from ESRGAN~\cite{ledig2017photo}, photo-realistic IR becomes prevalent due to the increasing requirement for high-quality image generation. Early research explored a variety of methods that combine GANs~\cite{goodfellow2020generative,mirza2014conditional} and other perceptual losses~\cite{johnson2016perceptual,zhang2018unreasonable,ding2020image} to train networks to predict images following the natural image distribution~\cite{ledig2017photo,wang2018esrgan,wang2021real}. However, GAN-based approaches often suffer from unstable performance and can be challenging to train on small datasets. Recent works therefore introduce diffusion models in image restoration for realistic image generation~\cite{li2022srdiff,saharia2022image,kawar2022denoising,luo2023image,luo2023refusion,ozdenizci2023restoring}. Moreover, leveraging pretrained Stable Diffusion (SD) models~\cite{rombach2022high,podell2023sdxl} as the prior is growing popular in real-world and blind IR tasks~\cite{lin2023diffbir,wang2023exploiting,wu2023seesr,yu2024scaling}. In particular, StableSR~\cite{wang2023exploiting} and DiffBIR~\cite{lin2023diffbir} adapt the SD model to image restoration using an approach similar to ControlNet~\cite{zhang2023adding}.  CoSeR~\cite{sun2023coser}, SeeSR~\cite{wu2023seesr}, and SUPIR~\cite{yu2024scaling} further introduce the textual semantic guidance in diffusion models for more accurate restoration performance.


%% file: sec/3_method.tex
\begin{figure*}[ht]
\begin{center}
\includegraphics[width=.95\linewidth]{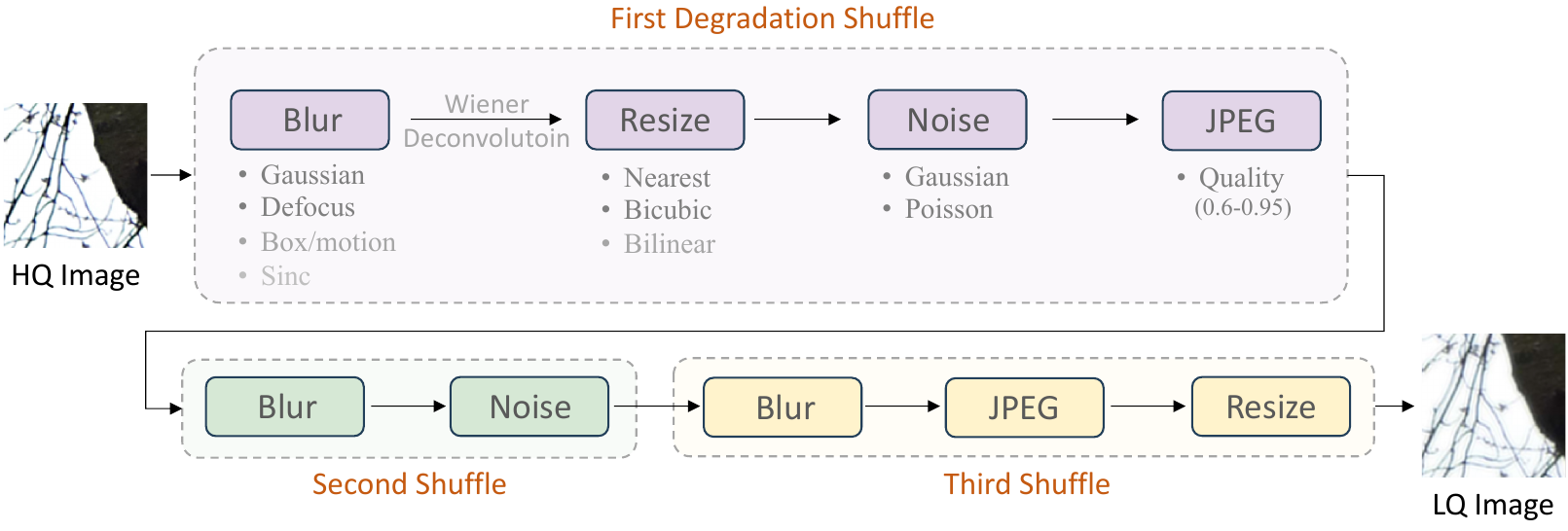}\vspace{-3.0mm}
\end{center}
    \caption{Overview of the proposed pipeline for synthetic image degradation. There are three degradation phases adopting the random shuffle strategy. We use different types of filters in blur generation and add the Wiener deconvolution for simulating ringing artifacts similar to the Sinc filter in Real-ESRGAN~\cite{wang2021real}. As a general $\times 1$ image restoration pipeline, we use one `resize' operation to provide image resolution augmentation, and another resize operation to ensure that all the degraded images are resized back to their original size.}
\label{fig:deg-pipeline}
\end{figure*}

\section{Method}
\label{sec:method}
Our work is a set of extensions and improvements on the degradation-aware CLIP \cite{luo2023controlling} which, in turn, builds on a mean-reverting SDE \citep{luo2023image}. Thus, before going into our contributions in the following sections, we first summarize the main constructions of the mean-reverting SDE and degradation-aware CLIP.

\subsection{Preliminaries}
\label{sec:preliminaries}

\paragraph{Mean-Reverting SDE}
Given a random variable $x_0$ sampled from an unknown distribution, $x_0 \sim p_0(x)$, the mean-reverting SDE~\cite{luo2023image} is defined according to
\begin{equation}
	dx = \theta_t \, (\mu - x) d t + \sigma_t d w, 
	\label{equ:ou}
\end{equation}
where $\theta_t$ and $\sigma_t$ are predefined time-dependent coefficients and $w$ is a standard Wiener process. By restricting the coefficients to satisfy $\sigma_t^2 \, / \, \theta_t = 2 \, \lambda^2$ for all timesteps $t$, we can solve the marginal distribution $p_t(x)$ as follows:
\begin{equation}
\begin{split}
    p_t(x) & \ = \mathcal{N}\bigl(x_t \cond m_t, v_{t}\bigr),\\
    m_t &= \mu + (x_0 - \mu) \, \expp^{-\bar{\theta}_{t}}, \\
    v_{t} & = \lambda^2 \, \Bigl(1 - \expp^{-2 \, \bar{\theta}_{t}}\Bigr),
\end{split}\label{eq:sde_solution}
\end{equation}
where $\bar{\theta}_{t} = \int^t_0 \theta_z \diff z$. To simulate the image degradation process, we set the HQ image as the initial state $x_0$ and the LQ image as the mean $\mu$. Then the forward SDE iteratively transforms the HQ image into the LQ image with additional noise, where the noise level is fixed to~$\lambda$. 

Moreover, \citet{anderson1982reverse} states that the forward process (\cref{equ:ou}) has a reverse-time representation as
\begin{equation}
    dx = \big[ \theta_t \, (\mu - x) - \sigma_t^2 \, \nabla_{x} \log p_t(x) \big] d t + \sigma_t d \hat{w},
    \label{eq:reverse-irsde}
\end{equation}
where $\nabla_{x}\log p_t(x)$ is the score function, which can be computed via~\cref{eq:sde_solution} during training since we have access to the ground truth LQ-HQ pairs in the training dataset. Following IR-SDE~\cite{luo2023image}, we train the score prediction network with a maximum likelihood loss which specifies the optimal reverse path $x_{t-1}^*$ for all times:
\begin{equation}
\begin{split}
    {x}_{t-1}^{*} &= \frac{1 - \expp^{-2 \, \bar{\theta}_{t-1}}}{1 - \expp^{-2 \, \bar{\theta}_t}} \expp^{-\theta_t^{'}} ({x}_t - \mu) \\[.6em]
    &\quad+ \frac{1 - \expp^{-2 \, \theta_t^{'}}}{1 - \expp^{-2 \, \bar{\theta}_t}} \expp^{-\bar{\theta}_{t-1}} (x_0 - \mu) + \mu,
\end{split}\label{eq:mll_loss}
\end{equation}
where $\theta_i^{'} = \int_{i-1}^i \theta_t dt$. The proof can be found in~\cite{luo2023image}. Once trained, we can simulate the backward SDE (\cref{eq:reverse-irsde}) to restore the HQ image, similar to what is done in other diffusion-based models~\cite{song2020score}.

\paragraph{Degradation-Aware CLIP} 
The core component of the degradation-aware CLIP (DACLIP~\cite{luo2023controlling}) is a controller that explicitly classifies degradation types and, more importantly, adapts the fixed CLIP image encoder~\cite{radford2021learning} to output high-quality content embeddings from corrupted inputs for accurate multi-task image restoration. DACLIP uses a contrastive loss to optimize the controller. Moreover, the training dataset is constructed with image-caption-degradation pairs where all captions are obtained using BLIP~\cite{li2022blip} on the clean HQ images of a multi-task dataset.

The trained DACLIP model is then applied to downstream networks to facilitate image restoration. Specifically, the cross-attention~\cite{rombach2022high} mechanism is introduced to incorporate image content embeddings to learn semantic guidance from the pre-trained DACLIP. For the unified image restoration task, the predicted degradation embedding is useful and can be combined with visual prompt learning~\cite{zhou2022learning} modules to further improve the performance.

\subsection{Synthetic Image Degradation Pipeline}
\begin{figure}[ht]
\begin{center}
\includegraphics[width=1.\linewidth]{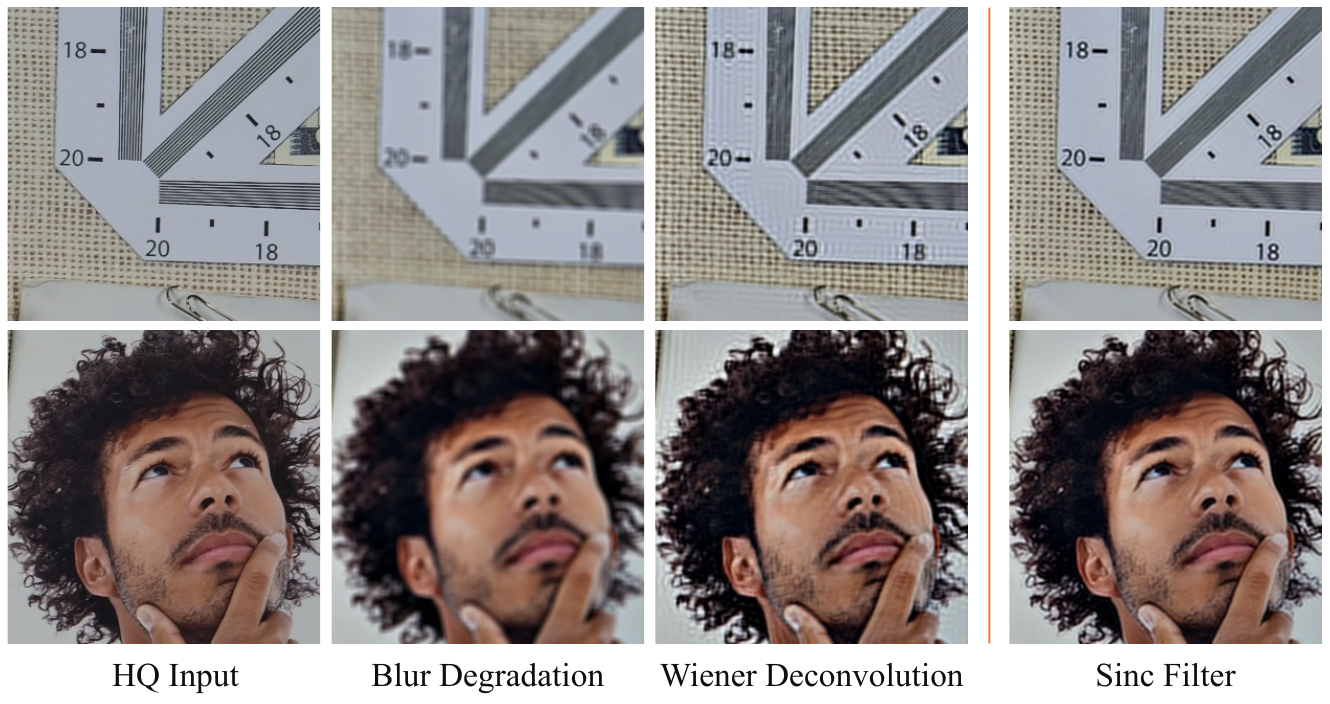}\vspace{-3.0mm}
\end{center}
    \caption{Examples of applying Wiener deconvolution to generate ringing artifacts. Compared to the Sinc filter used in Real-ESRGAN~\cite{wang2021real}, the proposed Wiener deconvolution generates more distinct ringing artifacts on textures.}
\label{fig:wiener-example}
\end{figure}

To restore clean images from unknown and complex degradations, we use a synthetic image degradation pipeline for LQ image generation, as shown in \cref{fig:deg-pipeline}. Common degradation models like \textbf{blur}, \textbf{resize}, \textbf{noise}, and \textbf{JPEG} compression are repeatedly involved to simulate complex scenarios. Following the \textit{high-order degradation} in Real-ESRGAN~\cite{wang2021real}, all degradation models in our pipeline have individual parameters that are randomly picked in each training step, which substantially improves the generalization for out-of-distribution datasets~\cite{wang2021real,zhang2021designing,luo2022deep}. In particular, in the blur model, we add some specific filter types (e.g., defocus, box, and motion filters) rather than only Gaussian filters for more general degradations, and the Wiener deconvolution is included to simulate natural ringing artifacts (which usually occurs in the preprocessing steps in some electronic cameras~\cite{yuan2007image,kupyn2018deblurgan}). Wiener deconvolution generates more distinct ringing artifacts on textures than the Sinc filter~\cite{wang2021real}, which can be seen in the two examples of applying Wiener deconvolution to blurry images in~\cref{fig:wiener-example}. 
For $\times 1$ image restoration (no resolution changes), we use two resize operations (with different interpolation modes) to provide random resolution augmentation and ensuring that all degraded images then are resized back to their original size. Note that our model focuses on image restoration in the wild (wild IR) and we therefore set all degradations to be light and diverse. Moreover, we randomly shuffle the degradation orders to further improve the generalization.

\subsection{Robust Degradation-Aware CLIP}
As introduced in \cref{sec:preliminaries}, DACLIP leverages a large-scale pretrained vision-language model, namely CLIP, for multi-task image restoration. While it works well on some (relatively) large and distinct degradation types such as rain, snow, shadow, inpainting, etc., it fares worse on the wild IR task since most degradations are mild, hard to describe in text, and contain multiple degradations in the same image. 

\begin{figure}[t]
\begin{center}
\includegraphics[width=1.\linewidth]{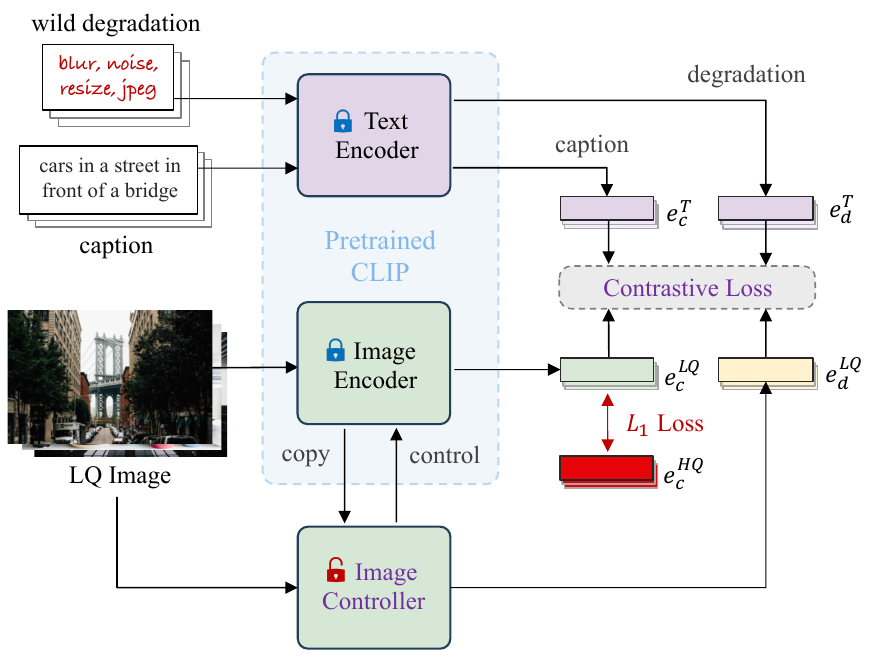}\vspace{-3.0mm}
\end{center}
    \caption{The proposed robust degradation-aware CLIP (DACLIP) model. $e^T_c$ and $e^T_d$ are caption and degradation text embeddings, respectively. The embeddings $(e^{LQ}_c$, $e^{LQ}_d)$ are extracted from LQ images, and $e^{HQ}_c$ represents the HQ image embedding extracted from the original CLIP image encoder.}
\label{fig:robust-daclip}
\end{figure}

To address this problem, we update DACLIP to learn more robust embeddings with the following aspects: 1) In dataset construction, instead of only using one degradation for each image, we use different combinations of degradation types such as \textit{`an image with blur, noise, ringing artifacts'} as the degradation text. 2) We add an $\ell_1$ loss to minimize the embedding distance between LQ and HQ images, where the HQ image embedding is extracted from the frozen CLIP image encoder. An overview of the robust DACLIP is illustrated in \cref{fig:robust-daclip}. The multi-degradation texts enable DACLIP to handle images that contain multiple complex degradations in the wild. Moreover, the additional $\ell_1$ loss forces DACLIP to learn accurate clean embeddings from our synthetic corrupted inputs. 

As \citet{luo2023controlling} illustrates, the quality of the image content embedding significantly affects the restoration results, thus encouraging us to extend the DACLIP base encoders to larger models for better performance. Specifically, we first generate clean captions using HQ images and then train the ViT-L-14 (rather than ViT-B-32 in DACLIP) based on the synthetic image-caption-degradation pairs, where the LQ images are generated following the pipeline in \cref{fig:deg-pipeline}. The dimensions of both image and text embeddings have increased from 512 to 768, which introduces more details for downstream IR models. 

We use IR-SDE~\cite{luo2023image} for realistic image generation and insert the image content embedding into the U-Net via cross-attention~\cite{rombach2022high}, analogously to what was done in \cite{luo2023controlling}. Since the degradation level is difficult to describe using text (e.g., the blurry level, noise level, and quality compression rate), we thus abandon the use of degradation embeddings for wild image restoration in both training and testing, similar to the single task setting in DACLIP~\cite{luo2023controlling}. In addition, to enable large-size inputs, we simply modify the network with an additional downsampling layer and an upsampling layer before and after the U-Net for model efficiency.

\subsection{Optimal Posterior Sampling for IR-SDE}
It is worth noting that the forward SDE in \cref{equ:ou} requires many timesteps to converge to a relatively stable state, i.e. a noisy LQ image with noise level $\lambda$. The sampling process (HQ image generation) uses the same timesteps as the forward SDE and is also sensitive to the noise scheduler~\cite{nichol2021improved}. To improve the sample efficiency, \citet{zhang2024entropy} propose a posterior sampling approach by specifying the optimal mean and variance in the reverse process. However, their method sets the SDE mean $\mu$ to 0, and only uses it to generate actions as a typical diffusion policy in reinforcement learning applications. In this work, we extend their posterior sampling strategy into a more general form for IR-SDE.

Let us use the same notation as in \cref{sec:preliminaries}. Formally, given the initial state $x_0$ and any other diffusion state $x_t$ at time $t \in [1,T]$, we can prove that the posterior of the mean-reverting SDE is tractable when conditioned on $x_0$. More specifically, this posterior distribution is given by
\begin{equation}
    p(x_{t-1} \cond x_t, x_0) = \mathcal{N}(x_{t-1} \cond \tilde{\mu}_t(x_t, \, x_0), \; \tilde{\beta}_t I),
\label{eq:posterior}
\end{equation}
which is a Gaussian with mean and variance given by:
\begin{equation}
\begin{split}
    \tilde{\mu}_t(x_t, x_0) &= \frac{1 - \expp^{-2 \, \bar{\theta}_{t-1}}}{1 - \expp^{-2 \, \bar{\theta}_t}} \expp^{-\theta_t^{'}} ({x}_t - \mu) \\[.6em]
    &\quad+ \frac{1 - \expp^{-2 \, \theta_t^{'}}}{1 - \expp^{-2 \, \bar{\theta}_t}} \expp^{-\bar{\theta}_{t-1}} (x_0 - \mu) + \mu,
\end{split}\label{eq:posterior_mu_var}
\end{equation}
\begin{equation}
    \mathrm{and} \quad \tilde{\beta}_t = \frac{(1 - {\expp^{-2\bar{\theta}_{t-1}}})(1 - {\expp^{-2\theta^{'}_t}})}{1 - \expp^{-2\bar{\theta}_{t}}}. \quad \quad
\end{equation}
Note that the posterior mean $\tilde{\mu}_t(x_t, x_0)$ has exactly the same form as the optimal reverse path $x^*_{t-1}$ in \cref{eq:mll_loss}, meaning that sampling from this posterior distribution is also optimal for recovering the initial state, i.e. the HQ image.

In addition, combining the reparameterization trick ($x_t = m_t + \sqrt{v_t} \, \epsilon_t$) with the noise prediction network $\tilde{\epsilon}_{\phi}(x_t, \mu, t)$ gives us a simple way to estimate $x_0$ at time~$t$:
\begin{equation}
    \hat{x}_0
    = \expp^{\bar{\theta}_{t}} 
        \big (x_t - \mu - \sqrt{v_t} \tilde{\epsilon}_{\phi}(x_t, \mu, t)
        \big) + \mu,
    \label{eq:est_x0}
\end{equation}
where $m_t$ and $v_t$ are the forward mean and variance in \cref{eq:sde_solution}, and $\phi$ is the learnable parameters. Then we iteratively sample reverse states based on this posterior distribution starting from noisy LQ images for efficient restoration.


%% file: sec/4_experiment.tex
\section{Experiments}
\label{sec:experiment}

We provide evaluations on different image restoration tasks to illustrate the effectiveness of the proposed method. 

\paragraph{Implementation Details}
For all experiments, we use the AdamW~\cite{loshchilov2017decoupled} optimizer with $\beta_1=0.9$ and $\beta_2=0.99$. The initial learning rate is set to $2 \times 10^{-4}$ and decayed to 1e-6 by the Cosine scheduler for 500\thinspace000 iterations. The noise level is fixed to 50 and the number of diffusion denoising steps is set to 100 for all tasks. We set the batch size to 16 and the training patches to 256 $\times$ 256 pixels. All models are implemented with PyTorch~\cite{paszke2019pytorch} and trained on a single A100 GPU for about 3-4 days.

\begin{figure*}[t]
  \centering
  \begin{minipage}{1.\linewidth}
    \centering
    \includegraphics[width=.99\linewidth]{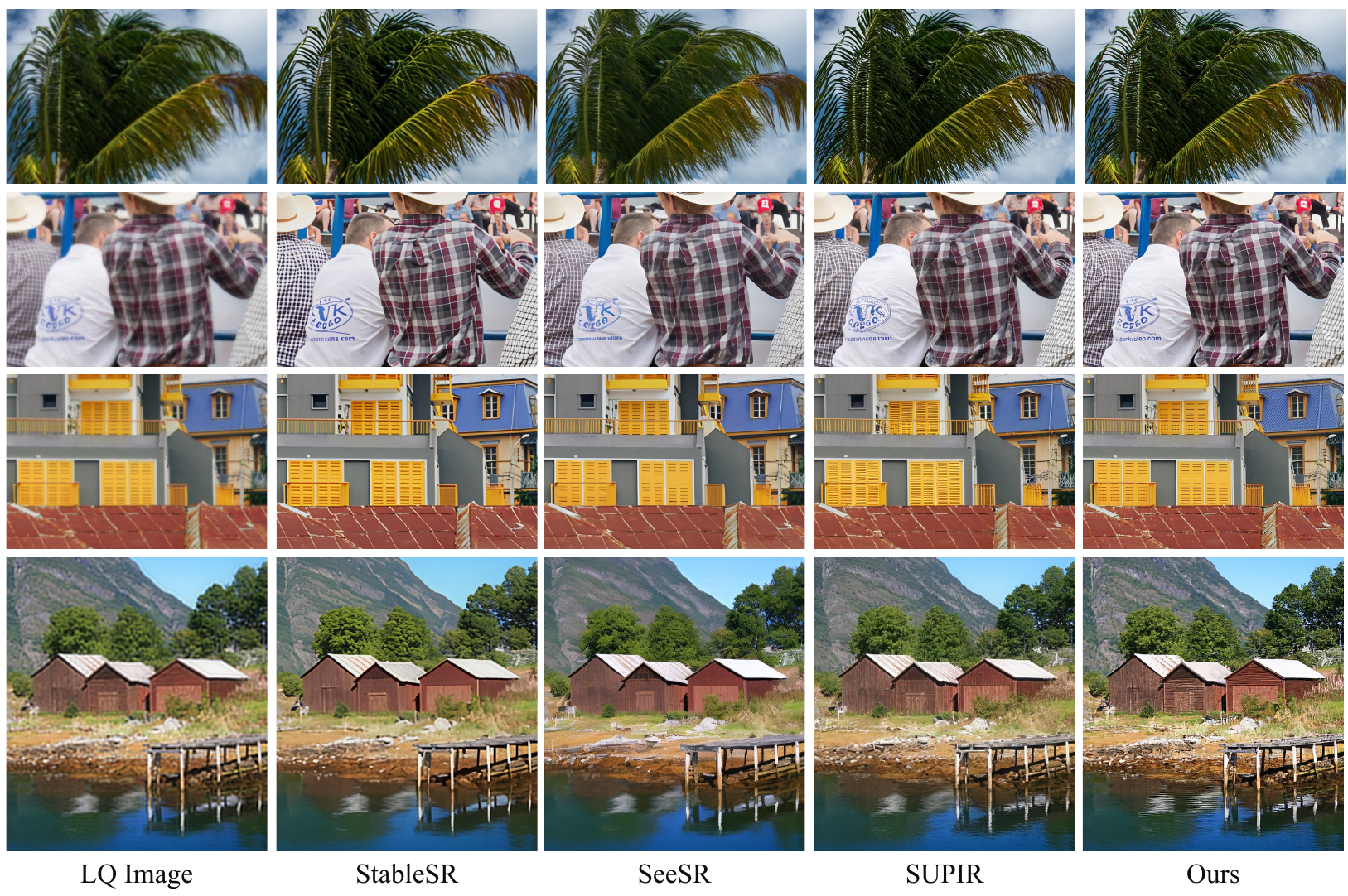}
    \vspace{-0.1in}
    \caption{Visual comparison of the proposed model with other state-of-the-art photo-realistic image restoration approaches on our synthetic DIV2K~\cite{agustsson2017ntire} dataset. Our method trains the diffusion model from scratch while other approaches leverage pretrained Stable Diffusion models. Note that all methods using Stable Diffusion are prone to generate unrecognizable text, such as for the white shirt in the second row.}
    \label{fig:div2k}
  \end{minipage}
  \vskip 0.15in
  \begin{minipage}{1.\linewidth}
    \centering
    \includegraphics[width=.99\linewidth]{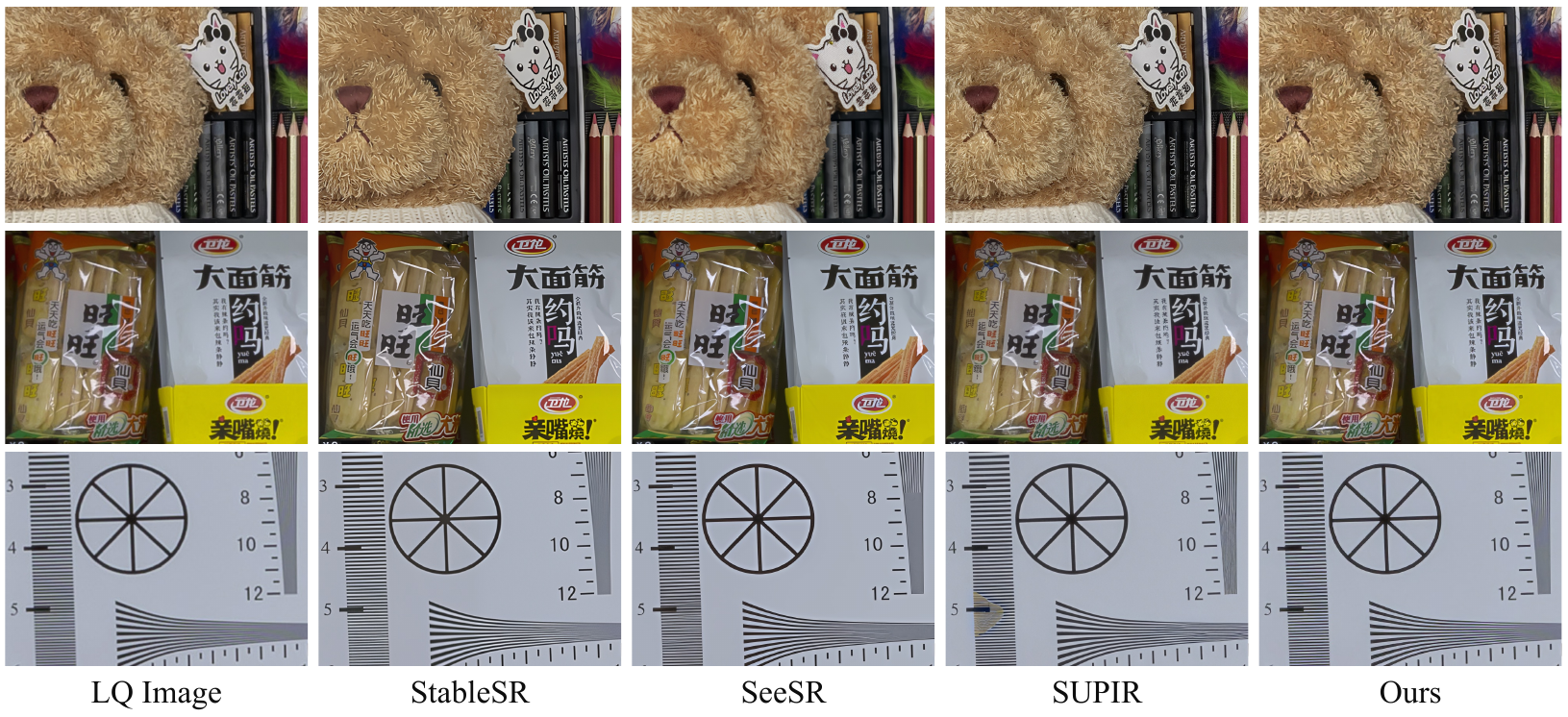}
    \vspace{-0.1in}
    \caption{Visual comparison of the proposed model with other state-of-the-art photo-realistic image restoration approaches on the RealSR $\times 2$~\cite{cai2019toward} dataset. Our method trains the diffusion model from scratch while other approaches leverage pretrained Stable Diffusion models.}
    \label{fig:realsr}
  \end{minipage}
\end{figure*}

\subsection{Evaluation of IR in the Wild}
\label{sec:exp_wildir}

\paragraph{Datasets and Metrics}
We train our model on the LSDIR dataset~\cite{li2023lsdir} which contains 84\thinspace991 high-quality images with rich textures and their downsampled versions. In training, we only utilize the collected HQ images and synthetically generate all HQ-LQ image pairs following the proposed degradation pipeline in \cref{fig:deg-pipeline}. In testing, we evaluate our model on two external datasets: DIV2K~\cite{agustsson2017ntire} and RealSR $\times 2$~\cite{cai2019toward}. Specifically, the DIV2K dataset contains 100 2K resolution image pairs with all LQ images generated using our degradation pipeline, while the RealSR $\times 2$ dataset contains 30 high-resolution real-world captured image pairs. In both datasets, we upscale all LQ images to have the same size as the corresponding HQ images for $\times 1$ image restoration.
For wild IR, we pay more attention to the visual quality of restored images and thus prefer to compare perceptual metrics such as LPIPS~\cite{zhang2018unreasonable}, DISTS~\cite{ding2020image}, FID~\cite{heusel2017gans}, and NIQE~\cite{mittal2012making}. Note that NIQE is a non-reference metric that only evaluates the quality of the output. In addition, we also report distortion metrics like PSNR and SSIM since we also want the prediction to be consistent with the input.

\paragraph{Comparison Approaches}
We compare our method DACLIP-IR with other state-of-the-art photo-realistic wild image restoration approaches: Real-ESRGAN~\cite{wang2021real}, StableSR~\cite{wang2023exploiting}, SeeSR~\cite{wu2023seesr}, and SUPIR~\cite{yu2024scaling}. All these comparison methods use the same degradation pipeline as that in Real-ESRGAN. Moreover, StableSR, SeeSR, and SUPIR employ pretrained Stable Diffusion models~\cite{rombach2022high,podell2023sdxl} as diffusion priors for better generalization on out-of-distribution images. SeeSR and SUPIR further leverage powerful vision-language models (RAM~\cite{zhang2023recognize} and LLaVA~\cite{liu2024visual}, respectively) to provide additional textual prompt guidance for image restoration in the wild.

\paragraph{Results} 
The quantitative results on the DIV2K and RealSR $\times 2$ datasets are summarized in \Cref{table:div2k} and \Cref{table:realsr}, respectively. It is observed that DACLIP-IR achieves the best performance over all approaches on the two datasets. The results are quite expected for the DIV2k dataset since we use the same degradation pipeline in both training and testing. For the RealSR $\times 2$ images, their degradations are unseen for all approaches and our DACLIP-IR still outperforms other methods on most metrics. Moreover, one can observe that changing the degradation pipeline directly decreases the performance on both datasets. And it is worth noting our SDE model is trained from scratch while all other diffusion-based approaches (StableSR, SeeSR, and SUPIR) leverage pretrained Stable Diffusion models as priors, demonstrating the effectiveness of the proposed method and our new degradation pipeline. 

A visual comparison of the proposed method with other state-of-the-art photo-realistic IR approaches on the two datasets is illustrated in \cref{fig:div2k} and \cref{fig:realsr}. One can see that all these methods can restore visually high-quality images. Moreover, results produced by SeeSR and SUPIR seem to have more details than StableSR, indicating the importance of textual guidance in diffusion-based image restoration. But in terms of distortion metrics that measure the consistency w.r.t the input, we found that pretrained Stable Diffusion models might introduce unclear priors and thus tend to generate text stroke adhesion which is unrecognizable, for example on the back of the white shirt in the second-row of~\cref{fig:div2k}. And in some cases, the SUPIR further produces fake textures and block artifacts, as shown in the third-row of~\cref{fig:div2k} (the yellow window frames) and the third-row of~\cref{fig:realsr} (the weird block around `5'). Although our method trains the diffusion model from scratch, its results still look realistic and are consistent with the inputs.

\begin{table}[t]
\caption{Quantitative comparison between the proposed method with other real-world image restoration approaches on our synthetic DIV2K~\cite{agustsson2017ntire} test set. `$^\dag$' means that our model is trained with the Real-ESRGAN~\cite{wang2021real} degradation pipeline.}
\label{table:div2k}
\vspace{-0.2in}
\begin{center}
\resizebox{1.\linewidth}{!}{
\begin{tabular}{lcccccc}
\toprule
\multirow{2}{*}{Method} &  \multicolumn{2}{c}{Distortion} & \multicolumn{4}{c}{Perceptual}  \\ \cmidrule(lr){2-3} \cmidrule(lr){4-7}
&  PSNR$\uparrow$ & SSIM$\uparrow$ & LPIPS$\downarrow$ & DISTS$\downarrow$ & FID$\downarrow$ & NIQE$\downarrow$  \\
\midrule
Real-ESRGAN~\cite{wang2021real}  & 27.71 & 0.810 & 0.200 & 0.107 & 27.32 & 4.41  \\
StableSR~\cite{wang2023exploiting}  & 26.04 & 0.759 & 0.241 & 0.123 & 34.74 & 4.11  \\
SeeSR~\cite{wu2023seesr}  & 26.29 & 0.721 & 0.223 & 0.114 & 27.94 & 3.56  \\
SUPIR~\cite{yu2024scaling}  & 26.81 & 0.741 & 0.194 & 0.099 & 21.73 & 3.52  \\

DACLIP-IR$^\dag$  & 27.56 & 0.796 & 0.195 & 0.113 & 24.32 & 3.43  \\

DACLIP-IR (Ours)  & \textbf{29.93} & \textbf{0.837} & \textbf{0.153} & \textbf{0.085} & \textbf{15.94} & \textbf{3.24} \\

\bottomrule
\end{tabular}
}
\end{center}
\vskip -0.1in
\end{table}

\begin{table}[t]
\caption{Quantitative comparison between the proposed method with other real-world IR approaches on the RealSR $\times 2$~\cite{cai2019toward} test set. All inputs are pre-upsampled with scale factor 2. `$^\dag$' means our model trained with the Real-ESRGAN~\cite{wang2021real} degradation pipeline.}
\label{table:realsr}
\vspace{-0.2in}
\begin{center}
\resizebox{1.\linewidth}{!}{
\begin{tabular}{lcccccc}
\toprule
\multirow{2}{*}{Method} &  \multicolumn{2}{c}{Distortion} & \multicolumn{4}{c}{Perceptual}  \\ \cmidrule(lr){2-3} \cmidrule(lr){4-7}
&  PSNR$\uparrow$ & SSIM$\uparrow$ & LPIPS$\downarrow$ & DISTS$\downarrow$ & FID$\downarrow$ & NIQE$\downarrow$  \\
\midrule
Real-ESRGAN~\cite{wang2021real}  & 28.03 & 0.855 & 0.151 & 0.117 & 47.65 & 4.84  \\
StableSR~\cite{wang2023exploiting}  & 27.55 & 0.838 & 0.169 & \textbf{0.112} & 54.87 & 5.45  \\
SeeSR~\cite{wu2023seesr}  & 28.38 & 0.815 & 0.212 & 0.139 & 40.85 & 4.20  \\
SUPIR~\cite{yu2024scaling}  & 29.32 & 0.826 & 0.175 & 0.122 & 31.75 & 4.61  \\

DACLIP-IR$^\dag$  & 28.92 & 0.858 & 0.184 & 0.138 & 33.76 & \textbf{4.19}  \\

DACLIP-IR (Ours)  & \textbf{30.65} & \textbf{0.878} & \textbf{0.148} & 0.113 & \textbf{30.09} & 4.31 \\

\bottomrule
\end{tabular}
}
\end{center}
\end{table}

\paragraph{Results on the NTIRE Challenge}

We also evaluate our model on the NTIRE 2024 `Restore Any Image Model (RAIM) in the Wild' challenge~\cite{liang2024ntire}, as shown in~\Cref{table:challenge}. To generalize to the challenge dataset, we first train our model on LSDIR~\cite{li2023lsdir} with synthetic image pairs, and then fine-tune it on a mixed dataset that contains both synthetic and real-world images from LSDIR~\cite{li2023lsdir} and RealSR~\cite{cai2019toward}. Note that we use the same model for both phase two and phase three of the challenge, but employ the original reverse-time SDE sampling in phase three for better visual results (small noise makes the photo look more realistic).

\begin{table}[t]
\renewcommand{\arraystretch}{1.}
\caption{Final results of the NTIRE 2024 RAIM challenge.}
\label{table:challenge}
\vspace{-0.1in}
\centering
\resizebox{.79\linewidth}{!}{
\begin{tabular}{lcccc}
\toprule
Team    &  Phase 2       & Phase 3    &   Final Score     & Rank     \\ \midrule
MiAlgo & 79.13	& 57 & 91.65 & 1 \\
Xhs-IAG & 81.96	& 47 & 82.07 & 2 \\
So Elegant & 79.69	& 46 & 80.09 & 3 \\
IIP IR & 80.03	& 14 & 45.94 & 4 \\
\textbf{DACLIP-IR} & 78.65	& 9 & 40.03 & 5 \\ \hline
TongJi-IPOE & 72.99	& 11 & 39.91 & 6 \\
ImagePhoneix & 78.93 & 4 & 34.79 & 7 \\
HIT-IIL & 69.80 & 1 & 27.92 & 8 \\

\bottomrule
\end{tabular}}
\end{table}

\begin{figure}[t]
\begin{center}
\includegraphics[width=0.925\linewidth]{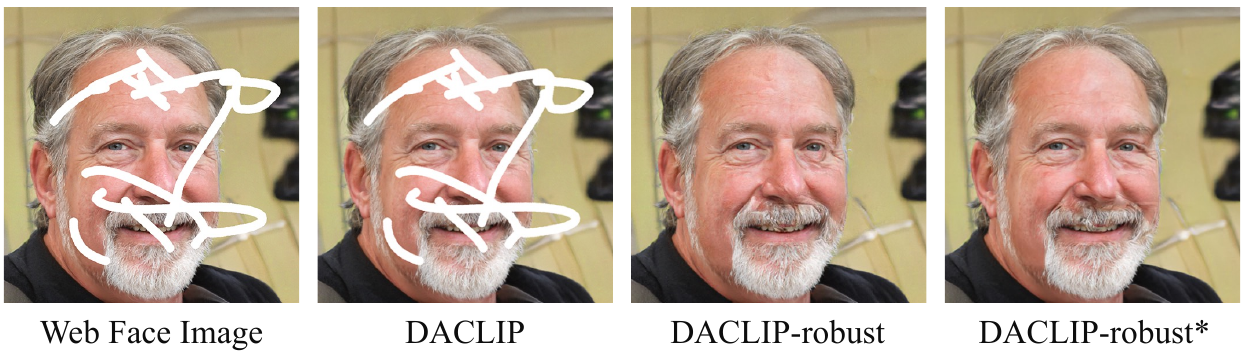}\vspace{-4.0mm}
\end{center}
    \caption{Inpainting results on a web-downloaded face image.}
\label{fig:face}
\end{figure}

\begin{table*}[t]
\begin{center}
\caption{Comparison of different methods on the unified image restoration task. `robust' means we add mild synthetic degradations (e.g., resize, noise, and JPEG) to LQ images in training as a data augmentation strategy for out-of-distribution data generalization. `*' means the method uses the proposed optimal posterior sampling approach for image generation. Here we report the results on the RainDrop~\cite{qian2018attentive}, LOL~\cite{wei2018deep}, and CBSD68~\cite{martin2001database} datasets for raindrop removal, low-light enhancement, and denoising task evaluation, respectively.}\vspace{-2.5mm}
\label{table:posterior}
\begin{small}
\resizebox{1.\linewidth}{!}{
\begin{tabular}{lcccccccccccc}
\toprule
\multirow{2}{*}{Method} &  \multicolumn{4}{c}{RainDrop~\cite{qian2018attentive}} & \multicolumn{4}{c}{LOL~\cite{wei2018deep}} & \multicolumn{4}{c}{CBSD68~\cite{martin2001database}} \\ \cmidrule(lr){2-5} \cmidrule(lr){6-9} \cmidrule(lr){10-13}
&  PSNR$\uparrow$ & SSIM$\uparrow$ & LPIPS$\downarrow$ & FID$\downarrow$ &  PSNR$\uparrow$ & SSIM$\uparrow$ & LPIPS$\downarrow$ & FID$\downarrow$ &  PSNR$\uparrow$ & SSIM$\uparrow$ & LPIPS$\downarrow$ & FID$\downarrow$  \\
\midrule
AirNet~\cite{li2022all}  & 30.68 & 0.926 & 0.095 & 52.71 & 14.24 & 0.781 & 0.321 & 154.2 & 27.51 & 0.769 & 0.264 & 93.89 \\
PromptIR~\cite{potlapalli2023promptir}  & 31.35 & \textbf{0.931} & 0.078 & 44.48 & \textbf{23.14} & 0.829 & 0.140 & 67.15 & \textbf{27.56} & \textbf{0.774} & 0.230 & 84.51 \\
IR-SDE~\cite{luo2023image}  & 28.49 & 0.822 & 0.113 & 50.22 & 16.07 & 0.719 & 0.185 & 66.42 & 24.82 & 0.640 & 0.232 & 79.38 \\
DACLIP~\cite{luo2023controlling}  & 30.81 & 0.882 & 0.068 & 38.91 & 22.09 & 0.796 & 0.114 & 52.23 & 24.36 & 0.579 & 0.272 & 64.71 \\
\midrule
DACLIP-robust  & 30.82 & 0.869 & 0.078 & 27.96 & 22.05 & 0.782 & 0.136 & 51.01 & 23.90 & 0.543 & 0.310 & 74.83 \\
DACLIP-robust*  & \textbf{31.68} & 0.921 & \textbf{0.051} & \textbf{21.92} & 22.78 & \textbf{0.848} & \textbf{0.092} & \textbf{41.50} & 25.86 & 0.723 & \textbf{0.167} & \textbf{62.12} \\

\bottomrule
\end{tabular}
}
\end{small}
\end{center}
\vskip -0.1in
\end{table*}

\begin{figure*}[ht]
\centering
\includegraphics[width=1.\linewidth]{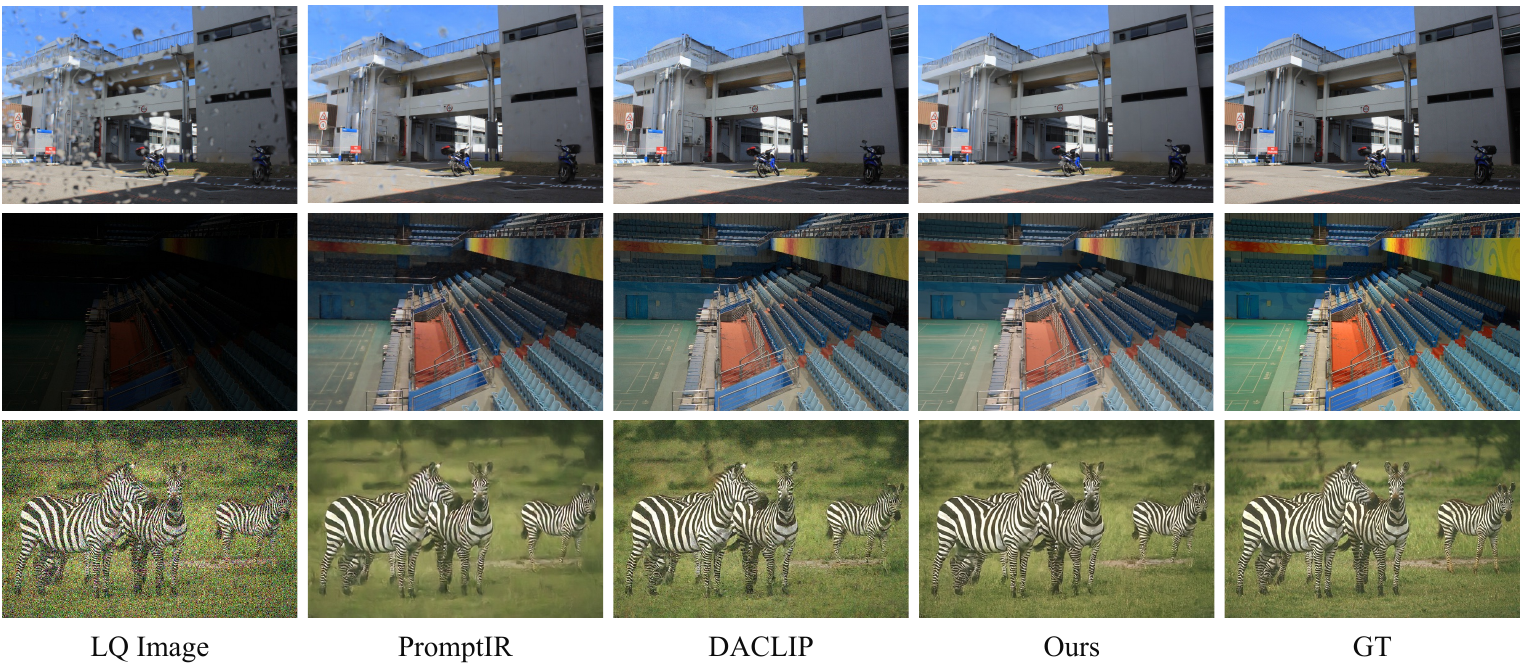}\vspace{-2.0mm}
\caption{Visual comparison of the proposed posterior sampling for the DACLIP model on the unified IR task.}
\label{fig:posterior_results}
\end{figure*}

\subsection{Effectiveness of the Posterior Sampling}
\label{sec:exp_posterior}

This section adopts the same settings as the DACLIP~\cite{luo2023controlling} and focuses on unified image restoration (UIR) which trains and evaluates a single model on multiple IR tasks. 

\paragraph{Robust DACLIP Model}
Notice that the original DACLIP is sensitive to input degradations since it is trained on specific datasets without data augmentation. To address this issue, we follow the synthetic training idea from wild image restoration and propose a robust DACLIP model. Similar to the original DACLIP, this robust model is trained on 10 datasets for unified image restoration. However, we now also add mild degradations such as noise, resize, and JPEG compression (first part of the degradation pipeline in~\cref{fig:deg-pipeline}) to the LQ images for data augmentation. The resulting model can then better handle real-world inputs that contain minor corruptions. \cref{fig:face} shows a face inpainting comparison for a web-downloaded image example. As one can see, the original DACLIP model completely fails to inpaint this image. On the other hand, the robust DACLIP model restores the face well, illustrating its robustness.

\paragraph{Evaluation and Analysis}
To analyze the effectiveness of the proposed posterior sampling, we choose 3 (out of 10) tasks for evaluation: raindrop removal on the RainDrop~\cite{qian2018attentive} dataset, low-light enhancement on the LOL~\cite{wei2018deep} dataset, and color image denoising on the CBSD68~\cite{martin2001database} dataset. The comparison methods include recent all-in-one image restoration approaches: AirNet~\cite{li2022all}, PromptIR~\cite{potlapalli2023promptir}, IR-SDE~\cite{luo2023image}, and the original DACLIP~\cite{luo2023controlling}. Finally, the posterior sampling is applied to our robust DACLIP model.
The comparison results are reported in \Cref{table:posterior}.  The PromptIR performs better on distortion metrics (PSNR and SSIM) while diffusion-based approaches have better perceptual performance (LPIPS and FID). Although the robust DACLIP model involves more degradations in training, it still performs similarly to its original version. By using the proposed posterior sampling in inference, the performance of the robust DACLIP model is significantly improved across all metrics and tasks. Especially for the denoising task, posterior sampling leads to the best LPIPS and FID performance, proving its effectiveness.




%% file: sec/5_conclusion.tex
\section{Conclusion}
\label{sec:conclusion}

This paper addresses the problem of photo-realistic image restoration in the wild. Specifically, we present a new degradation pipeline to generate low-quality images for synthetic data training. This pipeline includes diverse degradations (e.g., different blur kernels) and a random shuffle strategy to increase the generalization. Moreover, we improve the degradation-aware CLIP by adding multiple degradations to the same image and minimizing the embedding distance between LQ-HQ image pairs to enhance the LQ image embedding. Subsequently, we present a posterior sampling approach for IR-SDE, which significantly improves the performance of unified image restoration. Finally, we evaluate our model on various tasks and the NTIRE RAIM challenge and the results demonstrate that the proposed method is effective for image restoration in the wild.